\pdfoutput=1

\documentclass[11pt]{article}

\usepackage{EMNLP2022}

\usepackage{times}
\usepackage{latexsym}

\usepackage[T1]{fontenc}

\usepackage[utf8]{inputenc}

\usepackage{microtype}

\usepackage{inconsolata}

\usepackage{microtype}
\usepackage{graphicx}
\usepackage{amsmath}
\usepackage{amssymb}
\usepackage{booktabs}
\usepackage{multirow}
\usepackage{subcaption}
\usepackage{booktabs, multirow} 
\usepackage{soul}

\def\nytimes{NYTimes800k}
\def\goodnews{GoodNews}

\def\spacy{spaCy}
\def\joganic{JoGANIC}

\title{Focus! Relevant and Sufficient Context Selection \\for News Image Captioning}

\author{
Mingyang Zhou$^1$ \quad Grace Luo$^3$ \quad  Anna Rohrbach$^3$ \quad  Zhou Yu$^2$  
\\ 
$^1$University of California, Davis   \quad  $^2$ Columbia University  \\  $^3$University of California, Berkeley \\
minzhou@ucdavis.edu, zy2461@columbia.edu, \{graceluo, anna.rohrbach\}@berkeley.edu \\
}

\begin{document}
\maketitle
\begin{abstract}
News Image Captioning requires describing an image by leveraging additional context from a news article. Previous works only coarsely leverage the article to extract the necessary context, which makes it challenging for models to identify relevant events and named entities. In our paper, we first demonstrate that by combining more fine-grained context that captures the key named entities (obtained via an oracle) and the global context that summarizes the news, we can dramatically improve the model's ability to generate accurate news captions. This begs the question, how to automatically extract such key entities from an image? We propose to use the pre-trained vision and language retrieval model CLIP to localize the visually grounded entities in the news article and then capture the non-visual entities via an open relation extraction model. Our experiments demonstrate that by simply selecting a better context from the article, we can significantly improve the performance of existing models and achieve new state-of-the-art performance on multiple benchmarks. 

\end{abstract}

\section{Introduction}

\begin{figure*}[!ht]
\begin{scriptsize}
\begin{center}
\includegraphics[width=0.85\linewidth]{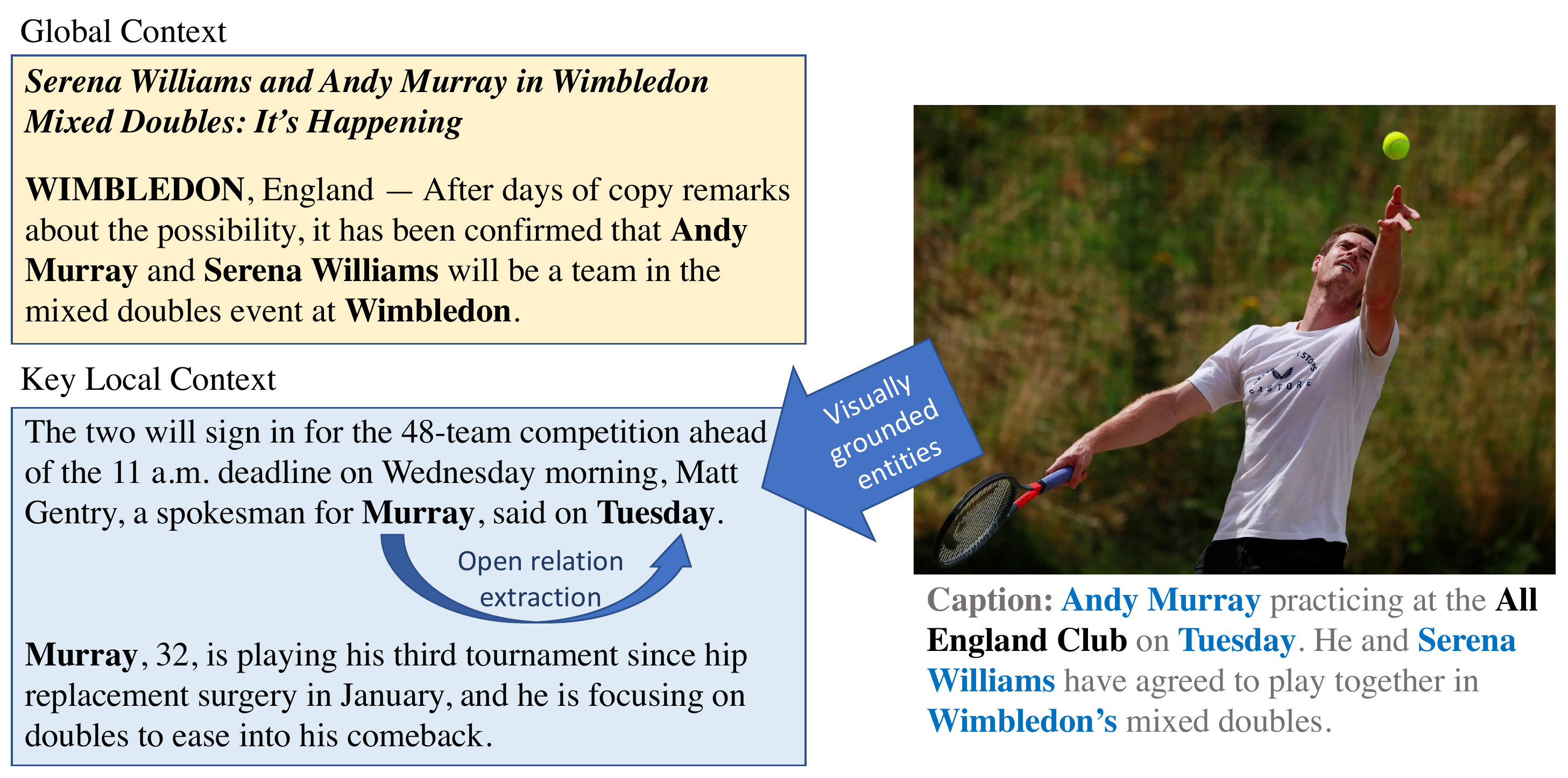}
\end{center}
\caption{We propose to use a combination of key local context and global context extracted from the news article. Both contain useful complementary information. While automatically extracting the key local context, we (a) detect visually grounded entities (e.g. Murray) and (b) discover non-visual entities via open relation extraction (e.g. Tuesday). Still, some caption entities (e.g. All
England Club) may be absent in the article.}
\label{fig:teaser}
\end{scriptsize}
\end{figure*}

News Image Captioning is an extension of the standard image captioning task, where, besides the image, one needs to leverage longer context in the form of the news article. This is important in order to capture the global news story context to properly discuss the image. At the same time, the specific entities (e.g. people) in the image are also often referenced in the article. 

Coincidentally, generating the named entities (names, locations, dates, etc), is one of the most critical challenges in news image captioning. 
We analyze the captions and the corresponding articles on three popular benchmarks, GoodNews~\cite{Biten_2019_CVPR}, NY800KTimes~\cite{Tran_2020_CVPR},  VisualNews~\cite{liu2020visualnews}, and find that 58.7$\%$ -- 74.5$\%$ of the named entities present in captions also appear in the articles.
This shows that the key information of a news caption, namely the named entities, can often be directly derived from the associated news article. However, due to the length and abundance of other information in the full articles, it is challenging to uncover the key entities. Inspired by this, we explore how to select relevant yet sufficient context from the news article to assist with the news image caption generation.

To analyze what the right type of context is, we start with an oracle-based study where we demonstrate that combining key local context and global context leads to the best result. Next, we move to designing a multimodal retrieval method to handle automatic extraction of key local context (key named entities). Our method consists of two stages, first we retrieve the visually grounded entities, then we discover the non-visual ones via open relation extraction. Figure~\ref{fig:teaser} illustrates our approach. Our full method outperforms the vanilla baseline on three benchmarks, and achieves the new state-of-the-art results on two of them.

\section{Related Work}
Automatic image captioning has achieved tremendous success over the past few years, where many methods\cite{showtell, densecap, KarpathyL15} are developed to accurately describe the visual objects and their relationships in the image. Recently, news image captioning has started to get increased attention with the focus on producing narration of news images with richer human-like information derived from the news article. The research on news image captioning is inspired by several recently collected benchmark datasets including BreakingNews~\cite{breakingnews}, GoodNews~\cite{Biten_2019_CVPR}, NY800KTimes~\cite{Tran_2020_CVPR}, and VisualNews~\cite{liu2020visualnews}. Earlier works such as \citet{breakingnews} and \citet{Biten_2019_CVPR} propose a two-stage pipeline, where they first generate template captions and then insert named entities into corresponding positions. \citet{INCEP} propose a hierarchical article encoding mechanism, where the model first retrieves the most relevant sentences and then attends to the appropriate words in the retrieved context with images involved in both steps. Later works, like  \citet{liu2020visualnews} and \citet{Tran_2020_CVPR} introduce end-to-end models that directly encode articles and images with pre-trained transformer architectures (BERT~\cite{devlin-etal-2019-bert} and RoBERTa~\cite{Liu2019RoBERTaAR}), then generate the caption directly. The end-to-end method is better than the template-based approach as it models the rich contextual information in named entities using byte-pair encoding (BPE)~\cite{BPE}. A recent work by \citet{yang-etal-2021-journalistic} proposes to follow journalistic guidelines when generating news image captions, i.e. each caption must contain the \textit{who}, \textit{when}, \textit{where}, and \textit{how} entities. 

There is rarely any effort spent on context selection to help news caption generation in previous work. The LSTM-based method such as \citet{breakingnews} and \citet{Biten_2019_CVPR} coarsely encode the full article either as a general embedding vector or a set of sentence embedding vectors. Later, the transformer-based approach such as \citet{Tran_2020_CVPR} and \citet{liu2020visualnews} directly encode the article at the token-level and achieved much better performance than the LSTM-based approach. However, they usually need to truncate the article to meet the encoding length limitation of the transformer architecture, which inevitably leads to the loss of useful information. \citet{INCEP} is one of few works that addresses context selection for news image caption, where they train a cross-modal retrieval model VSE++ \cite{faghri2018vse++} to find the sentences from the article that are most relevant to the image. However, the retrieved relevant sentences from VSE++ are limited to only capture visually grounded information. We argue that other valuable information, i.e. non-visual named entities, should be included as the relevant context to help news image caption generation. 


\begin{table*}[!ht]\centering
\small
\begin{tabular}{l|c|c|c|cc}\toprule
\multirow{2}{*}{ Context } &\multirow{2}{*}{BLEU-4} &\multirow{2}{*}{ROUGE} &\multirow{2}{*}{CIDER} & \multicolumn{2}{c}{Named Entities} \\
& & & &P &R \\\cmidrule{1-6}
Original &6.0 &21.4 &53.8 &22.2 &18.7 \\
\midrule
Oracle Key Local (Par) &6.9 &23.0 &62.7 &29.1 &23.8 \\
\midrule
Oracle Key Local (Sent) &7.1 &23.5 &66.4 &\textbf{31.9} &\textbf{25.6} \\
\midrule
Oracle Key Local (Sent) + Global &\textbf{7.2} &\textbf{24.2} &\textbf{67.4} &30.0 &24.5  \\
\bottomrule
\end{tabular}
\caption{Evaluation results on GoodNews with different strategies for selecting the article context. The Original context here means the first 500 words from the article following \cite{Tran_2020_CVPR}. 
}
\label{tab:goodnews_study}
\end{table*}

\begin{table*}[!ht]\centering
\small
\begin{tabular}{l|c|c|c|cc}\toprule
\multirow{2}{*}{ Context } &\multirow{2}{*}{BLEU-4} &\multirow{2}{*}{ROUGE} &\multirow{2}{*}{CIDER} & \multicolumn{2}{c}{Named Entities} \\
& & & &P &R \\\cmidrule{1-6}
Original &6.3 &21.7 &54.4 &24.6 &22.2 \\
\midrule
Oracle Key Local (Par) &9.3 &26.4 &81.4 &38.7 &34.7 \\
\midrule
Oracle Key Local (Sent) &9.7 &27.1 &83.9 &\textbf{41.9} &\textbf{37.2} \\
\midrule
Oracle Key Local (Sent) + Global &\textbf{10.3} &\textbf{27.6} &\textbf{84.5} &39.8 &35.5  \\
\bottomrule
\end{tabular}
\caption{Evaluation results on NYTimes800K with different strategies for selecting the article context. The Original context here means the nearest 512 tokens surrounding the image following \cite{Tran_2020_CVPR}.}
\label{tab:nytimes_study}
\end{table*}



\section{Oracle Study: Optimal News Context Selection}
\label{sec:optimal_context}
We posit that there are two critical types of context in the news article that we need to extract to help the model generate accurate news image captions: (1) \textbf{The key local context}: a snippet of the news article that contains the named entities mentioned in the caption. (2) \textbf{The global context} that summarizes the news story. We propose a strategy to extract these two types of context using an oracle (ground-truth caption) and validate their contribution by training a model using them as input. We compare against the models trained on the news article with alternative schemes of context selection. Next, we introduce the details of the datasets, captioning model, and evaluation metrics that we use for our study to validate our hypothesis on optimal news context selection. 

\subsection{Datasets} 
\label{sec:truncated contex}
We use two well-known news image captioning datasets: \goodnews~\cite{Biten_2019_CVPR} and \nytimes~\cite{Tran_2020_CVPR}. We follow the data split setting in \citet{Tran_2020_CVPR} with 421K training, 18K validation, and 23K testing for the \goodnews{} and 763K training, 8K validation, and 22K test for the \nytimes. 

\subsection{News Captioning Model} 
We conduct our experiment with \emph{Transform and Tell}~\cite{Tran_2020_CVPR} (\emph{Tell} for short), which is a state-of-the-art transformer-based captioning model. It encodes the article with a pre-trained RoBERTa model \cite{Liu2019RoBERTaAR} and extracts the representation of an image via a set of visual encoders including ResNet152~\cite{ResNet}, YOLOv3 object detector~\cite{Yolov3}, and MTCNN face detector~\cite{MCTNN} to represent different visual concepts. Then the extracted multi-modal feature is passed to a transformer decoder to generate the caption. 

Due to the maximum length limitation of encoded text handled by RoBERTa, the full article is truncated before feeding into the language encoder. \citet{Tran_2020_CVPR} apply two different strategies to truncate the article. For \goodnews, they simply choose the first 500 words.
For \nytimes, as the image position in the original news article is available, they select the 512 tokens ``surrounding'' the image, which may provide the model with more image-relevant context. 

\subsection{Evaluation Metrics} 
Following previous literature~\cite{Tran_2020_CVPR, liu2020visualnews, yang-etal-2021-journalistic}, we use BLEU-4 \cite{papineni-etal-2002-bleu}, ROUGE \cite{lin-2004-rouge}, and CIDEr scores \cite{CIDER} to measure the similarity between the generated caption and the referenced ground-truth captions. Among the three metrics, CIDEr is the most suitable one for news image captioning evaluation as it focuses more on unusual words generation. In addition, we evaluate the precision and
recall on named entities to verify whether the key information is covered in the generated caption. Specifically, we check the exact string matches between the named entities detected in both the ground truth captions and the generated captions using SpaCy \footnote{https://spacy.io/}. 

\subsection{Oracle Optimal Context Selection}
We construct the key local context for a given image, assuming access to its ground truth caption. 

Given a set of named entities present in the caption, we define the key local context as the set of sentences (or paragraphs) that contain these named entities
The selected sentences (paragraphs) are concatenated following the original order in the news article. We specifically compare the key information at both the sentence-level and paragraph-level to measure the impact of the different degrees of condensation. For global context, we concatenate the article title and the first paragraph to summarize the high-level news story. When we combine the two types of context, the global context is always put before the local key information context, following the presentation in the original paragraph. The content in the local context that overlaps with the global context will be omitted. We cap the length of the combination of the global context and the local key information context to 500 words to stay within the maximum sequence length of the transformer architecture.   

\subsection{Results and Discussion} The results of training Tell with different contexts on \goodnews{} and \nytimes{} are summarized in Tables \ref{tab:goodnews_study} and \ref{tab:nytimes_study}. The Original context corresponds to the respective scheme used by \citet{Tran_2020_CVPR} on each dataset (as described above). 
It is clear that when we feed the model more relevant context that contains the key information, the performance of the model is dramatically improved on all the reference-based metrics. Meanwhile, the precision and recall of the named entities in the generated caption are significantly increased by over 5$\%$ 
in both \goodnews{} and \nytimes. We also observe that when the key information is condensed to an even more concise level (from paragraph to sentence),
the performance of the model can be further improved. This indicates that the model can learn more effectively from the condensed context that focuses on the critical information. 

When we combine the global context and the key local context, we find that we can achieve an improvement on the reference metrics but observe a drop on named entity accuracy in the generated captions. Adding global context inevitably adds more words which leads to the introduction of potentially irrelevant entities. However, it provides a valuable source of information to create captions that cover not only the named entities found in the image but also convey the key topic of the article. Therefore, we deem the combination of the global and local context as sufficient and relevant knowledge for the news captioning model.

\section{Automatic Key Local Context Detection}
\label{sec: auto_selection}
In the previous section, we identified an optimal strategy for selecting news context, which relied on the named entities found in the ground truth captions. In practice, automatically determining the key named entities remains an open challenge. To address it, we propose a simple multi-modal retrieval pipeline to extract the key named entities from the news article. The pipeline consists of two stages: 1. we identify visually grounded named entities, such as people's names and geographic locations, via cross-modal retrieval; 2. we discover the key named entities that are not visually grounded, e.g. time, by exploring factual relations between the entities in the article and the detected visually grounded entities. 

\begin{table}[!ht]\centering
\small
\begin{tabular}{@{}c|c|c@{}}\toprule
\spacy{} Type & Description & Components \\
\midrule
PERSON & People, including fictional & WHO \\
NORP & Political groups & WHO \\
ORG & Companies, agencies, etc & WHO \\
DATE & Dates or periods & WHEN \\
TIME & Times smaller than a day & WHEN \\
FAC & Buildings, airports, highways & WHERE \\
GPE & Countries, cities, states & WHERE \\
LOC & Locations, mountains, waters & WHERE \\
PRODUCT & Objects, vehicles, foods & MISC \\ 
EVENT & Named wars, sports events & MISC \\
ART & Titles of books, songs & MISC \\
LAW & Laws & MISC \\
LAN & Any named language & MISC \\
PERCENT & Percentage, including “$\%$” & MISC \\
MONEY & Monetary values & MISC \\
QUANTITY & Measurements & MISC \\
ORDINAL & “first”, “second”, etc & MISC \\
CARDINAL & Numerals & MISC \\
\bottomrule
\end{tabular}
\caption{Mapping from \spacy{} named entities types to the four named entities components defined in \cite{yang-etal-2021-journalistic}.}
\label{tab:ner_type_map}
\end{table}

\paragraph{Visually Grounded Named Entities Retrieval}
According to \citet{yang-etal-2021-journalistic}, the named entities in the news captions can be generally divided into four groups: WHO, WHEN, WHERE, and MISC. The mapping between the SpaCy named entity types and each category can be found in Table~\ref{tab:ner_type_map}. We define named entities belonging to WHO and WHERE as visually grounded named entities, as they are often related to specific image regions. For example, people's names can be associated to the faces and names of cities can be inferred from the landmarks shown in the image. It is also evident by the high ratio of appearance of these two entity types in the news captions \cite{yang-etal-2021-journalistic}. WHO appears in more than 93$\%$ and WHERE appears in more than 50$\%$ of the news captions in \goodnews{} and \nytimes. 

To localize the visually grounded named entities in the news article, we first find the relevant sentences from the entire news article that are semantically close to the image. Our approach is to leverage a large pre-trained cross-modal retrieval model CLIP~\cite{CLIP} trained on over 400 million image-text pairs from the Internet. The sentence relevance is measured by the cosine similarity with the image in the learned embedding space of CLIP. As there is a large domain discrepancy between news sentences and the training data of CLIP, we first need to fine-tune CLIP on the news image captioning dataset. The CLIP model is fine-tuned with a contrastive loss to distinguish the positive image-sentence pairs from the negative pairs. For the positive sentences we use the news image captions since there are no labels for which news article sentences are relevant to each image, and the writing style of the captions is close to that of the article sentences. Besides using captions describing other images as negative samples, we also define hard negatives as news article sentences that do not cover any non-stop words in the news caption. During image captioning, we use the trained CLIP model to rank news sentences against the image and pick the top 2 as the most relevant sentences. The visually grounded named entities are then retrieved as the WHO and WHERE entities contained in the retrieved relevant sentences. If the top 2 retrieved sentences do not contain any visually grounded entity, we keep searching the rest of the retrieved sentences descending by similarity score to the news image until at least a visually grounded entity is captured. 

\paragraph{Non-visually Grounded Named Entities Retrieval}
We define the named entities belonging to the rest of the categories: WHEN and MISC as non-visually grounded name entities. Such entities are not semantically related to the image and are more challenging to localize via cross-modal retrieval. However, they are often related to the visually grounded named entities in the news text. For example, in the sentence \textit{Ali Kashani-Rafye, started selling bastani (Persian for ice cream) in 1980 at his grocery.}, the TIME entity \textit{1980} is associated with the person's name \textit{Ali Kashani-Rafye} with the action \textit{selling bastani}.
The task of extracting such relations between named entities is known as open relation extraction. We leverage an open relation extraction method OpenNRE~\cite{han-etal-2019-opennre} to detect all the relations between each pair of named entities in every sentence of the news. We filter out the relations with a detection confidence score lower than 0.7. After filtering, if there is a relation extracted between a non-visually grounded named entity and a detected visually grounded named entity from the cross-modal retrieval, we use such non-visually grounded named entities during final context extraction. 

\paragraph{News Context Selection}
Once we get the full list of the detected named entities, we follow the optimal strategy derived in Sec \ref{sec:optimal_context} to select the  news context. We combine the local-focused context guided by the detected named entities and the global context that is consisted of the title and the first paragraph to format the relevant news context.  

\begin{table*}[!ht]\centering
\small
\begin{tabular}{l|c|c|c|cc}\toprule
\multirow{2}{*}{Model} &\multirow{2}{*}{BLEU-4} &\multirow{2}{*}{ROUGE} &\multirow{2}{*}{CIDER} & \multicolumn{2}{c}{Named Entities} \\
& & & &P &R \\\cmidrule{1-6}
VisualNews &6.1 &21.6 &55.4 &22.9 &19.3 \\
\joganic &\textbf{6.8} &\textbf{23.0} &\textbf{61.2} &\textbf{26.9} &\textbf{22.1}\\
\midrule
Tell (Original) &6.0 &21.4 &53.8 &22.2 &18.7  \\
Tell (Ours) &6.3 &22.4 &60.3 &24.2 &20.9 \\
\bottomrule
\end{tabular}
\caption{Comparison between Tell trained with our context selection and other SoTA methods on \goodnews.}
\label{tab:main_goodnews}
\end{table*}

\begin{table*}[!ht]\centering
\small
\begin{tabular}{l|c|c|c|cc}\toprule
\multirow{2}{*}{ Model } &\multirow{2}{*}{BLEU-4} &\multirow{2}{*}{ROUGE} &\multirow{2}{*}{CIDER} & \multicolumn{2}{c}{Named Entities} \\
& & & &P &R \\\cmidrule{1-6}
VisualNews &6.4 &21.9 &56.1 &24.8 &22.3 \\
\joganic &6.8 &22.8 &59.4 &28.6 &24.5\\
\midrule
Tell (Original) &6.3 &21.7 &54.4 &24.6 &22.2 \\
Tell (Ours) &\textbf{7.0} &\textbf{22.9} &\textbf{63.6} &\textbf{29.8} &\textbf{25.9} \\
\bottomrule
\end{tabular}
\caption{Comparison between Tell trained with our context selection and other SoTA methods on  \nytimes.}
\label{tab:main_nytimes}
\end{table*}

\section{Experiment}
In this section, we demonstrate the impact of context section for news image captioning in the automatic setting. We first introduce the baselines and the evaluation datasets. Then we discuss our findings from the experimental results. 

\subsection{Datasets}
We evaluate the performance of different methods on three well-known news image caption datasets, including \goodnews~\cite{Biten_2019_CVPR}, \nytimes~\cite{Tran_2020_CVPR}, and VisualNews~\cite{liu2020visualnews}. The details of \goodnews{} and \nytimes{} are introduced in Section~\ref{sec:optimal_context}. Unlike the other two datasets that are only collected from the New York Times, VisualNews is sourced from multiple news agencies, including The Guardian, BBC, USA Today, and The Washington Post. We follow \cite{liu2020visualnews} to split the data into 400K for training, 40k for validation, and 40k for testing.

\begin{table*}[!ht]\centering
\small
\begin{tabular}{l|c|c|c|cc}\toprule
\multirow{2}{*}{ Model } &\multirow{2}{*}{BLEU-4} &\multirow{2}{*}{ROUGE} &\multirow{2}{*}{CIDER} & \multicolumn{2}{c}{Named Entities} \\
& & & &P &R \\\cmidrule{1-6}
VisualNews &5.3 &17.9 &50.5 &19.7 &17.6 \\
\midrule
Tell (Original) &9.6 &22.8 &83.8 &23.7 &19.2 \\
Tell (Ours) &\textbf{11.6} &\textbf{25.0} &\textbf{107.6} &\textbf{26.2} &\textbf{21.2} \\
\bottomrule
\end{tabular}
\caption{Comparison between Tell trained with our context selection and other SoTA methods on  VisualNews.}
\label{tab:main_visualnews}
\end{table*}

\subsection{Compared Models}
\paragraph{VisualNews Captioner}~\cite{liu2020visualnews} is a transformer~\cite{attentionisall} based approach with several augmentations to enhance the named entity generation. They add a list of entities from the news article as additional input to guide the named entity generation in the captions. Meanwhile, they also propose to decode the out-of-vocabulary named entities as the entity type token instead of an unknown token. Then when an entity type token is decoded, they simply retrieve the most frequent named entity of the same type in the news article as the final generation. 
\paragraph{\joganic} \cite{yang-etal-2021-journalistic} proposes to generate news caption following journalistic principles where the caption must contain components such as \textit{who}, \textit{when}, \textit{where}, and \textit{how}. They propose to first generate the most likely caption components and then use the component-specific decoding method to generate the named entities that follow the guidance of the predicted components.
\paragraph{Tell (Original)} is the Transform and Tell model~\cite{Tran_2020_CVPR} trained on the truncated news articles. The truncation strategies for \goodnews{} and \nytimes{} were introduced in Section~\ref{sec:truncated contex}. For Visual News, following \goodnews, we also truncate the full article to 500 words. 

\paragraph{Tell (Ours)} is the Transform and Tell model \cite{Tran_2020_CVPR} trained on the context from our proposed automatic selection approach as described in Sec \ref{sec: auto_selection}.

\begin{table}[!ht]\centering
    \small
    \begin{tabular}{l|c|c|cc}\toprule
    \multirow{2}{*}{Context} &\multirow{2}{*}{BLEU-4} &\multirow{2}{*}{CIDER} & \multicolumn{2}{c}{Named Entities} \\
    & &  &P &R \\\cmidrule{1-5}
    LSTM (Original) &2.0 &13.9 &10.7&7.1  \\
    \midrule
    LSTM (Ours) &\textbf{2.1} &\textbf{15.4} &\textbf{11.4} &\textbf{7.6} \\
    \bottomrule
    \end{tabular}
\caption{Evaluation of an LSTM-based Captioner with our proposed context on NY800K.}
\label{tab:ablation_other_model}
\end{table}

\begin{table*}[!ht]\centering
    \small
    \begin{tabular}{l|c|c|c|cc}\toprule
    \multirow{2}{*}{Context} &\multirow{2}{*}{BLEU-4} &\multirow{2}{*}{ROUGE} &\multirow{2}{*}{CIDER} & \multicolumn{2}{c}{Named Entities} \\
    & & & &P &R \\\cmidrule{1-6}
    Tell (Original) &6.0 &21.8 &57.8 &22.5 &19.1  \\
    \midrule
    Tell (Only Visual NE) &5.7 &21.5 &54.4 &22.8 &19.2 \\
    \midrule
    Tell (Ours) &\textbf{6.3} &\textbf{22.4} &\textbf{60.3} &\textbf{24.2} &\textbf{20.9} \\
    \bottomrule
    \end{tabular}
\caption{Evaluation of Tell trained on the context guided by different sets of detected named entities on \goodnews{} including Visually Grounded Named Entities (Visual NE) and all the extracted named entities (our full approach).}
\label{tab:ablation_goodnews}
\end{table*}

\begin{table*}[!ht]\centering
\small
\begin{tabular}{l|c|c|c|cc}\toprule
\multirow{2}{*}{Context} &\multirow{2}{*}{BLEU-4} &\multirow{2}{*}{ROUGE} &\multirow{2}{*}{CIDER} & \multicolumn{2}{c}{Named Entities} \\
& & & &P &R \\\cmidrule{1-6}
Tell (Original) & 6.3 &21.7 &54.4 &24.6 &22.2 \\
\midrule
Tell (Only Visual NE) &6.8 &21.8 &59.4 &28.4 &23.2 \\
\midrule
Tell (Ours) &\textbf{7.0} & \textbf{22.9} & \textbf{63.6} &\textbf{29.8} & \textbf{25.9} \\
\bottomrule
\end{tabular}
\caption{Evaluation of Tell trained on the context guided by different sets of detected named entities on \nytimes{} including Visually Grounded Named Entities (Visual NE) and all the extracted named entities (our full approach).}
\label{tab:ablation_nytimes}
\end{table*}

\begin{table*}[!ht]\centering
\small
\begin{tabular}{l|c|c|c|cc}\toprule
\multirow{2}{*}{Context} &\multirow{2}{*}{BLEU-4} &\multirow{2}{*}{ROUGE} &\multirow{2}{*}{CIDER} & \multicolumn{2}{c}{Named Entities} \\
& & & &P &R \\\cmidrule{1-6}
Tell (Original) & 9.6 &22.8 &83.8 &23.7 &19.2 \\
\midrule
Tell (Only Visual NE) &10.9 &23.9 &96.4 &24.5 &20.7 \\
\midrule
Tell (Ours) &\textbf{11.6} &\textbf{25.0} &\textbf{107.6} &\textbf{26.2} &\textbf{21.2} \\
\bottomrule
\end{tabular}
\caption{Evaluation of Tell trained on the context guided by different sets of detected named entities on Visual News, including Visually Grounded Named Entities (Visual NE) and all the extracted named entities (our full approach).}
\label{tab:ablation_visualnews}
\end{table*}

\subsection {Results}
The results for the three benchmarks are summarized in Tables \ref{tab:main_goodnews}, \ref{tab:main_nytimes}, and \ref{tab:main_visualnews}. We observe that the Tell model trained on our proposed automatically selected context consistently outperforms the Tell model trained on the original context on all metrics in all three datasets. Meanwhile, Tell trained on our selected context also achieves the new state-of-the-art performance on \nytimes{} and Visual News. This demonstrates that even when the key named entities are automatically retrieved from the article, our proposed context selection strategy still helps the model learn to generate captions with better quality. We believe the main reason for the better performance is that the selected context covers more key named entities than the original context. 

We also notice that the Tell model trained with our selected context does not outperform \joganic{} on the \goodnews{} dataset. We think this is due to the lower coverage ratio of the caption-relevant named entities within the news articles compared to the other two datasets (58$\%$ vs. 70$\%$). This seems to limit somewhat the benefit of focusing the key context on the named entities, although we still improve over Tell (Original). Note, that we do outperform \joganic{} on the \nytimes{} dataset.

We also evaluate the usefulness of our proposed context selection strategy on different news captioning methods other than the transformer-based model Tell. We train an LSTM based news image captioning model\footnote{LSTM+GLOVE baseline in Transform and Tell \cite{Tran_2020_CVPR}} on our selected context and the original context with \nytimes{}. The results are summarized in Table \ref{tab:ablation_other_model}. It is clear that the LSTM captioner trained on the selected context is still consistently better, which demonstrates the generalization of our proposed context selection strategy to other news captioning approaches. 

\begin{figure*}[!ht]
\centering
\begin{subfigure}{\textwidth}
  \centering
  \includegraphics[width=\linewidth]{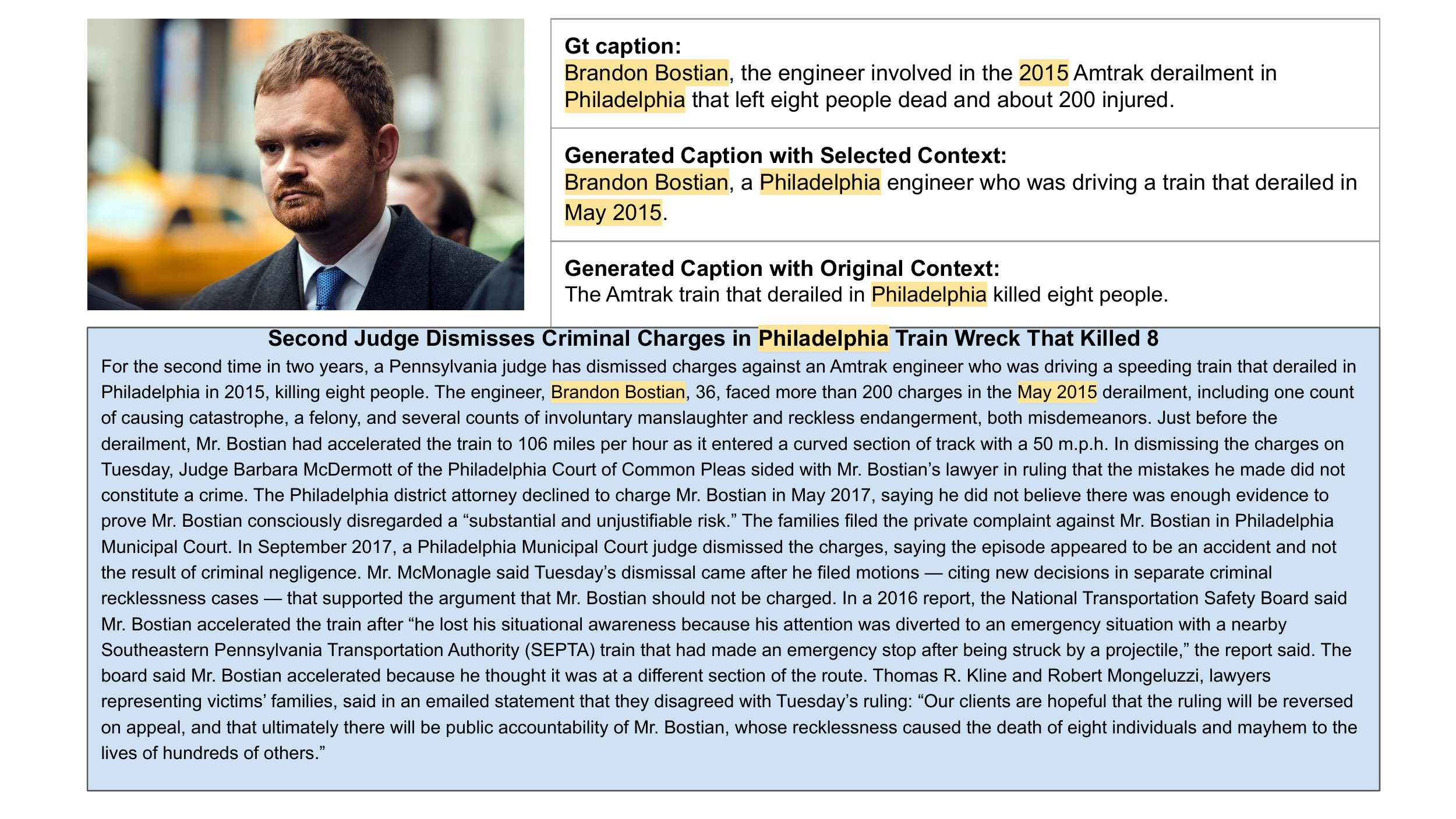}
  \caption{}
  \label{fig:sub1}
\end{subfigure} \hfill
\begin{subfigure}{\textwidth}
  \centering
  \includegraphics[width=\linewidth]{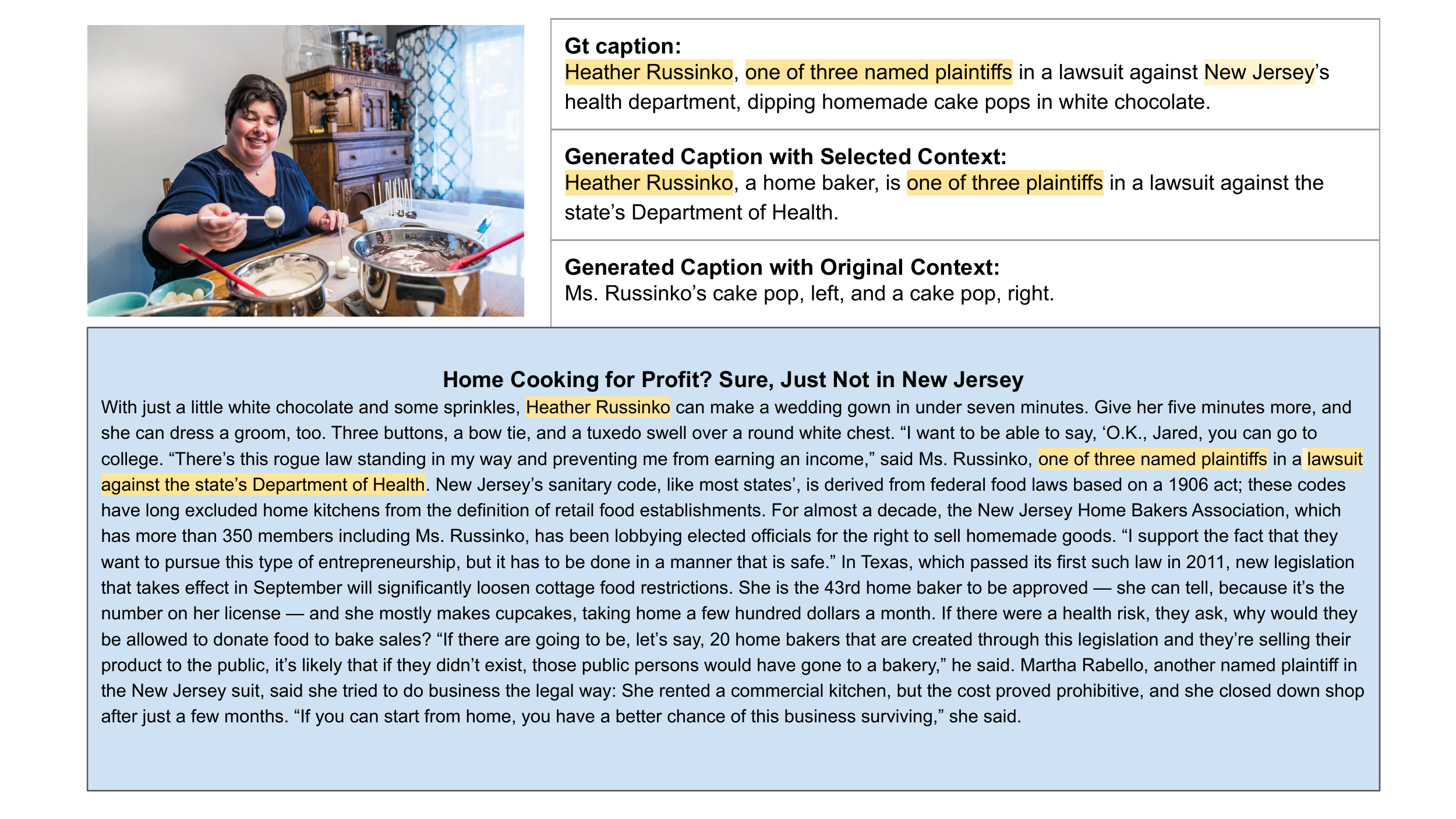}
  \caption{}
  \label{fig:sub2}
\end{subfigure}
\caption{Examples of caption generation. For each example, we display news image (top left block) and the corresponding captions (top right block) from the \nytimes{} test set. We compare the captions generated by the model trained on the original context and the model trained on our selected context to the ground truth caption. We also display the selected context used by our proposed strategy (bottom block). The key information in the generated captions, ground truth caption, and the selected context are highlighted with yellow bars.}
\label{fig:qualitative}
\end{figure*}

\subsection{Ablation on the Detected Named Entities}
We validate the contribution of the different named entities obtained by our proposed multi-modal retrieval pipeline. We compare using context guided only by the detected visually grounded named entities and that of using the full set of detected named entities. When we select the context with these sets of entities, we follow the optimal strategy introduced in Sec \ref{sec:optimal_context} to include both the key local context and the global context. We perform the ablation study on all three datasets, and the results are summarized in Tables~\ref{tab:ablation_goodnews}, \ref{tab:ablation_nytimes}, \ref{tab:ablation_visualnews}.

We observe that when we only extract the visually grounded named entities with the fine-tuned CLIP model, we can already select a better context to train the model than the original context on \nytimes{} and Visual News. However, the performance on \goodnews{} is still worse than that trained on the original context. This mixed effect can likely be explained by the significantly different average article length in the \goodnews{} vs. \nytimes{} and Visual News dataset. Since \goodnews{} often contains very short articles that can almost entirely be fit within the first 500 words, simply selecting context guided by the detected visually grounded named entities will lead to incomplete key information coverage. On the other hand, in \nytimes{} and Visual News, as the full articles contain much more than 500 words (on average above 900 words), when we truncate the context to 500 words models suffer from a great loss of useful information. Therefore, even when we simply detect the visually grounded named entities, we achieve a higher named entity coverage ratio than that of the Original Context. When we include the non-visually named entities detected via OpenNRE, we achieve additional improvement and outperform the model trained with the original context for all benchmarks. This demonstrates that non-visually named entities do add essential information for the model in order to generate captions with better quality.



\subsection{Qualitative Generation Results}
We also show qualitative examples that demonstrate the effectiveness of our proposed context selection strategy. In Fig \ref{fig:qualitative}, we show that the generated caption from the model trained on our selected context has covered more accurate key information than that trained on the original context. On the top example, the engineer's name ``\textit{Brandon Bostian}'' is captured by our multi-modal retrieval pipeline, and then our open relation extraction model successfully captures the relation between the TIME entity ``\textit{2015}'' with the detected name. These key pieces of information are correctly captured by our model that uses selected context, while they are missed by the model that uses the original unfiltered context. Meanwhile, the LOCATION named entity ``Philadelphia'' is captured by the global context (the title) and is also generated by the model trained on our selected context. Similarly, on the bottom, the caption generated from the model trained on our selected context contains the correct person entity (``Heather Russinko'') and key event (``one of three plaintiffs'') discussed in the article. 
\section{Conclusion}
Our paper explores news context selection to improve the performance of existing models on news caption generation. Our proposed strategy to select relevant context is two-fold:  (1) Include the key local context that focuses on the key information in the news caption, namely the named entities. (2) Include the global context, such as titles and the first paragraph, that summarizes the news story. Both parts are validated in an oracle setting with a strong news captioning model, Transform-and-Tell (Tell) \cite{Tran_2020_CVPR}, across multiple benchmarks. We further study key local context selection in a practical (automatic) setting, where we first detect a set of visual named entities with a multi-modal retrieval pipeline, find related non-visual named entities via relation extraction, then use these named entities to pick relevant news article sentences. The performance of Tell is consistently improved using our proposed context selection, and it achieves the new state-of-the-art on two benchmarks. We hope that our findings will inspire future work to develop models that even more effectively leverage the relevant context in the article to improve caption generation quality. Meanwhile, our proposed method can also be potentially extended to select a relevant context from external sources to cover the key named entities missing from the original news article. 
\begin{table}[!ht]\centering
    \small
    \begin{tabular}{l|c|c|cc}\toprule
    \multirow{2}{*}{Test Set} &\multirow{2}{*}{BLEU-4} &\multirow{2}{*}{CIDER} & \multicolumn{2}{c}{Named Entities} \\
    & &  &P &R \\\cmidrule{1-5}
    All &6.3 &60.3 &24.2 &20.9  \\
    \midrule
    High NE Cover &\textbf{7.3} &\textbf{65.1} &\textbf{31.2} &\textbf{28.2} \\
    \bottomrule
    \end{tabular}
\caption{Evaluation of Tell with our proposed context on the full Good News test set vs. the subset where the article contains more than 70\% of the ground truth caption's named entities.}
\label{tab:ablation_ne_cover}
\end{table}

\section{Limitations, Ethics, and Broader Impacts}
Our proposed context selection method performs a hard filtering strategy to keep just the context that focuses on the key named entities and short global context that summarizes the story, which inevitably might lead to losing some potentially useful information in the article. A more optimal strategy could be to develop a soft filtering technique where higher weights are assigned to the more relevant context and lower weights to the rest of the context during the encoding stage. 

We also find that our approach is more useful when there is a high ratio of covered named entities from the caption in the article. 
We compare the performance of Tell using our proposed context on the full test set vs. the subset of GoodNews where more than 70\% of caption-relevant named entities are mentioned in the article. The result is summarized in Table \ref{tab:ablation_ne_cover}. We find that the performance on this subset is much better than on the whole test set, where on average only articles only contain 59\% of the caption-relevant named entities.

Finally, since we leverage the large-scale pretrained CLIP~\cite{CLIP} model, we might transfer any biases (including gender or racial biases) that CLIP has learned into our context selection process. We recommend exercising caution when adopting our approach in practice.

\section*{Acknowledgements}
This work was supported in part by DARPA’s SemaFor and PTG programs, as well as BAIR’s industrial alliance programs.

\clearpage
\bibliography{anthology,custom}

\begin{thebibliography}{22}
\expandafter\ifx\csname natexlab\endcsname\relax\def\natexlab#1{#1}\fi

\bibitem[{Biten et~al.(2019)Biten, Gomez, Rusinol, and
  Karatzas}]{Biten_2019_CVPR}
Ali~Furkan Biten, Lluis Gomez, Marcal Rusinol, and Dimosthenis Karatzas. 2019.
\newblock Good news, everyone! context driven entity-aware captioning for news
  images.
\newblock In \emph{Proceedings of the IEEE/CVF Conference on Computer Vision
  and Pattern Recognition (CVPR)}.

\bibitem[{Devlin et~al.(2019)Devlin, Chang, Lee, and
  Toutanova}]{devlin-etal-2019-bert}
Jacob Devlin, Ming-Wei Chang, Kenton Lee, and Kristina Toutanova. 2019.
\newblock \href {https://doi.org/10.18653/v1/N19-1423} {{BERT}: Pre-training of
  deep bidirectional transformers for language understanding}.
\newblock In \emph{Proceedings of the 2019 Conference of the North {A}merican
  Chapter of the Association for Computational Linguistics: Human Language
  Technologies, Volume 1 (Long and Short Papers)}, pages 4171--4186,
  Minneapolis, Minnesota. Association for Computational Linguistics.

\bibitem[{Faghri et~al.(2018)Faghri, Fleet, Kiros, and
  Fidler}]{faghri2018vse++}
Fartash Faghri, David~J Fleet, Jamie~Ryan Kiros, and Sanja Fidler. 2018.
\newblock \href {https://github.com/fartashf/vsepp} {Vse++: Improving
  visual-semantic embeddings with hard negatives}.

\bibitem[{Han et~al.(2019)Han, Gao, Yao, Ye, Liu, and
  Sun}]{han-etal-2019-opennre}
Xu~Han, Tianyu Gao, Yuan Yao, Deming Ye, Zhiyuan Liu, and Maosong Sun. 2019.
\newblock \href {https://doi.org/10.18653/v1/D19-3029} {{O}pen{NRE}: An open
  and extensible toolkit for neural relation extraction}.
\newblock In \emph{Proceedings of EMNLP-IJCNLP: System Demonstrations}, pages
  169--174.

\bibitem[{He et~al.(2016)He, Zhang, Ren, and Sun}]{ResNet}
Kaiming He, Xiangyu Zhang, Shaoqing Ren, and Jian Sun. 2016.
\newblock \href {https://doi.org/10.1109/CVPR.2016.90} {Deep residual learning
  for image recognition}.
\newblock In \emph{2016 IEEE Conference on Computer Vision and Pattern
  Recognition (CVPR)}, pages 770--778.

\bibitem[{Hu et~al.(2020)Hu, Chen, and Jin}]{INCEP}
Anwen Hu, Shizhe Chen, and Qin Jin. 2020.
\newblock \href {https://doi.org/10.1145/3394171.3413576} {{ICECAP:}
  information concentrated entity-aware image captioning}.
\newblock In \emph{{MM} '20: The 28th {ACM} International Conference on
  Multimedia, Virtual Event / Seattle, WA, USA, October 12-16, 2020}, pages
  4217--4225. {ACM}.

\bibitem[{Johnson et~al.(2016)Johnson, Karpathy, and Fei-Fei}]{densecap}
Justin Johnson, Andrej Karpathy, and Li~Fei-Fei. 2016.
\newblock Densecap: Fully convolutional localization networks for dense
  captioning.
\newblock In \emph{Proceedings of the IEEE Conference on Computer Vision and
  Pattern Recognition}.

\bibitem[{Karpathy and Fei{-}Fei(2017)}]{KarpathyL15}
Andrej Karpathy and Li~Fei{-}Fei. 2017.
\newblock \href {https://doi.org/10.1109/TPAMI.2016.2598339} {Deep
  visual-semantic alignments for generating image descriptions}.
\newblock \emph{{IEEE} Trans. Pattern Anal. Mach. Intell.}, 39(4):664--676.

\bibitem[{Lin(2004)}]{lin-2004-rouge}
Chin-Yew Lin. 2004.
\newblock \href {https://aclanthology.org/W04-1013} {{ROUGE}: A package for
  automatic evaluation of summaries}.
\newblock In \emph{Text Summarization Branches Out}, pages 74--81, Barcelona,
  Spain. Association for Computational Linguistics.

\bibitem[{Liu et~al.(2020)Liu, Wang, Wang, and Ordonez}]{liu2020visualnews}
Fuxiao Liu, Yinghan Wang, Tianlu Wang, and Vicente Ordonez. 2020.
\newblock \href {http://arxiv.org/abs/2010.03743} {Visualnews : Benchmark and
  challenges in entity-aware image captioning}.

\bibitem[{Liu et~al.(2019)Liu, Ott, Goyal, Du, Joshi, Chen, Levy, Lewis,
  Zettlemoyer, and Stoyanov}]{Liu2019RoBERTaAR}
Y.~Liu, Myle Ott, Naman Goyal, Jingfei Du, Mandar Joshi, Danqi Chen, Omer Levy,
  M.~Lewis, Luke Zettlemoyer, and Veselin Stoyanov. 2019.
\newblock Roberta: A robustly optimized bert pretraining approach.
\newblock \emph{ArXiv}, abs/1907.11692.

\bibitem[{Papineni et~al.(2002)Papineni, Roukos, Ward, and
  Zhu}]{papineni-etal-2002-bleu}
Kishore Papineni, Salim Roukos, Todd Ward, and Wei-Jing Zhu. 2002.
\newblock \href {https://doi.org/10.3115/1073083.1073135} {{B}leu: a method for
  automatic evaluation of machine translation}.
\newblock In \emph{Proceedings of the 40th Annual Meeting of the Association
  for Computational Linguistics}, pages 311--318, Philadelphia, Pennsylvania,
  USA. Association for Computational Linguistics.

\bibitem[{Radford et~al.(2021)Radford, Kim, Hallacy, Ramesh, Goh, Agarwal,
  Sastry, Askell, Mishkin, Clark, Krueger, and Sutskever}]{CLIP}
Alec Radford, Jong~Wook Kim, Chris Hallacy, Aditya Ramesh, Gabriel Goh,
  Sandhini Agarwal, Girish Sastry, Amanda Askell, Pamela Mishkin, Jack Clark,
  Gretchen Krueger, and Ilya Sutskever. 2021.
\newblock \href {http://arxiv.org/abs/2103.00020} {Learning transferable visual
  models from natural language supervision}.
\newblock \emph{CoRR}, abs/2103.00020.

\bibitem[{Ramisa et~al.(2018)Ramisa, Yan, Moreno-Noguer, and
  Mikolajczyk}]{breakingnews}
Arnau Ramisa, Fei Yan, Francesc Moreno-Noguer, and Krystian Mikolajczyk. 2018.
\newblock \href {https://doi.org/10.1109/TPAMI.2017.2721945} {Breakingnews:
  Article annotation by image and text processing}.
\newblock \emph{IEEE Transactions on Pattern Analysis and Machine
  Intelligence}, 40(5):1072--1085.

\bibitem[{Redmon and Farhadi(2018)}]{Yolov3}
Joseph Redmon and Ali Farhadi. 2018.
\newblock \href {http://arxiv.org/abs/1804.02767} {Yolov3: An incremental
  improvement}.
\newblock \emph{CoRR}, abs/1804.02767.

\bibitem[{Sennrich et~al.(2016)Sennrich, Haddow, and Birch}]{BPE}
Rico Sennrich, Barry Haddow, and Alexandra Birch. 2016.
\newblock \href {https://doi.org/10.18653/v1/p16-1162} {Neural machine
  translation of rare words with subword units}.
\newblock In \emph{Proceedings of the 54th Annual Meeting of the Association
  for Computational Linguistics, {ACL} 2016, August 7-12, 2016, Berlin,
  Germany, Volume 1: Long Papers}. The Association for Computer Linguistics.

\bibitem[{Tran et~al.(2020)Tran, Mathews, and Xie}]{Tran_2020_CVPR}
Alasdair Tran, Alexander Mathews, and Lexing Xie. 2020.
\newblock Transform and tell: Entity-aware news image captioning.
\newblock In \emph{IEEE/CVF Conference on Computer Vision and Pattern
  Recognition (CVPR)}.

\bibitem[{Vaswani et~al.(2017)Vaswani, Shazeer, Parmar, Uszkoreit, Jones,
  Gomez, Kaiser, and Polosukhin}]{attentionisall}
Ashish Vaswani, Noam Shazeer, Niki Parmar, Jakob Uszkoreit, Llion Jones,
  Aidan~N Gomez, \L~ukasz Kaiser, and Illia Polosukhin. 2017.
\newblock \href
  {https://proceedings.neurips.cc/paper/2017/file/3f5ee243547dee91fbd053c1c4a845aa-Paper.pdf}
  {Attention is all you need}.
\newblock In \emph{Advances in Neural Information Processing Systems},
  volume~30. Curran Associates, Inc.

\bibitem[{Vedantam et~al.(2015)Vedantam, Zitnick, and Parikh}]{CIDER}
Ramakrishna Vedantam, C.~Lawrence Zitnick, and Devi Parikh. 2015.
\newblock \href {https://doi.org/10.1109/CVPR.2015.7299087} {Cider:
  Consensus-based image description evaluation}.
\newblock In \emph{2015 IEEE Conference on Computer Vision and Pattern
  Recognition (CVPR)}, pages 4566--4575.

\bibitem[{Xu et~al.(2015)Xu, Ba, Kiros, Cho, Courville, Salakhutdinov, Zemel,
  and Bengio}]{showtell}
Kelvin Xu, Jimmy~Lei Ba, Ryan Kiros, Kyunghyun Cho, Aaron Courville, Ruslan
  Salakhutdinov, Richard~S. Zemel, and Yoshua Bengio. 2015.
\newblock Show, attend and tell: Neural image caption generation with visual
  attention.
\newblock In \emph{Proceedings of the 32nd International Conference on
  International Conference on Machine Learning - Volume 37}, ICML'15, page
  2048–2057. JMLR.org.

\bibitem[{Yang et~al.(2021)Yang, Karaman, Tetreault, and
  Jaimes}]{yang-etal-2021-journalistic}
Xuewen Yang, Svebor Karaman, Joel Tetreault, and Alejandro Jaimes. 2021.
\newblock \href {https://doi.org/10.18653/v1/2021.emnlp-main.419} {Journalistic
  guidelines aware news image captioning}.
\newblock In \emph{Proceedings of the 2021 Conference on Empirical Methods in
  Natural Language Processing}, pages 5162--5175, Online and Punta Cana,
  Dominican Republic. Association for Computational Linguistics.

\bibitem[{Zhang et~al.(2016)Zhang, Zhang, Li, and Qiao}]{MCTNN}
K.~Zhang, Z.~Zhang, Z.~Li, and Y.~Qiao. 2016.
\newblock \href {https://doi.org/10.1109/LSP.2016.2603342} {Joint face
  detection and alignment using multitask cascaded convolutional networks}.
\newblock \emph{IEEE Signal Processing Letters}, 23(10):1499--1503.

\end{thebibliography}
\bibliographystyle{acl_natbib}

\clearpage

\appendix

\section{Context Selection vs CLIP Retrieval}
Previously, \cite{INCEP} demonstrate that cross-modal text retrieval model like VSE++ \cite{faghri2018vse++} can localize image relevant news context to help improve news caption generation quality. To compare our context strategy with the image-to-text retrieval context selection method, we use the fine-tuned CLIP model as a stronger image-to-text retrieval context and select the optimal context as the top-k sentences that has the highest similarity to the image based on CLIP's prediction. Following \cite{INCEP}, we iteratively sample k from 1 to 20 and find that the optimal value of k is 10 as it leads to the best CIDER score of Tell on the NY800K dataset. The comparison between the context selected by our proposed strategy and that retrieved by CLIP is summarised in Table \ref{tab:ablation_clip}. We observe that Tell trained on our selected context is consistently better than that is trained on CLIP retrieved news context. We find that this improvement comes from entities that cannot be captured by cross-modal retrieval, such as the non-visual named entities. CLIP-retrieved context also does not include the global context, which missed the high-level summary of the major story of the news. 

\begin{table}[!ht]\centering
    \small
    \begin{tabular}{l|c|c|cc}\toprule
    \multirow{2}{*}{Context} &\multirow{2}{*}{BLEU-4} &\multirow{2}{*}{CIDER} & \multicolumn{2}{c}{Named Entities} \\
    & &  &P &R \\\cmidrule{1-5}
    CLIP Retrieved &6.4 &57.5 &25.7 &22.7  \\
    \midrule
    Our &\textbf{7.0} &\textbf{63.6} &\textbf{29.8} &\textbf{25.9} \\
    \bottomrule
    \end{tabular}

\caption{Evaluation of Tell with our proposed context against Tell trained on CLIP retrieved top k sentences from news articles. }
\label{tab:ablation_clip}
\end{table}



\end{document}